\documentclass[letterpaper, 10 pt, conference]{ieeeconf}  

\IEEEoverridecommandlockouts                              

\usepackage{graphics} 
\usepackage{graphicx}
\usepackage{amsmath} 
\usepackage{multirow}
\usepackage{cite}
\usepackage[hidelinks]{hyperref}
\usepackage{orcidlink}

\title{\LARGE \bf Think Step by Step: Chain-of-Gesture Prompting for Error Detection in Robotic Surgical Videos
}

\author{Zhimin Shao\orcidlink{0000-0002-3078-0939}, Jialang Xu\orcidlink{0000-0003-2324-7033}, Danail Stoyanov\orcidlink{0000-0002-0980-3227}, \IEEEmembership{Fellow, IEEE}, \\ Evangelos B. Mazomenos\orcidlink{0000-0003-0357-5996}, \IEEEmembership{Member, IEEE} and Yueming Jin\orcidlink{0000-0003-3775-3877}, \IEEEmembership{Member, IEEE}
\thanks{This work has been submitted to the IEEE for possible publication. Copyright may be transferred without notice, after which this version may no longer be accessible.}
\thanks{This research was funded in whole, or in part, by Ministry of Education Tier 1 Start up grant, NUS, Singapore (A-8001267-01-00); Ministry of Education Tier 1 grant, NUS, Singapore (A-8001946-00-00); the Wellcome/EPSRC Centre for Interventional and Surgical Sciences (WEISS) [203145/Z/16/Z, NS/A000050/1]; the EPSRC-funded UCL Centre for Doctoral Training in Intelligent, Integrated Imaging in Healthcare (i4health) [EP/S021930/1]; a UCL Research Excellence Scholarship; the Department of Science, Innovation and Technology (DSIT); the Royal Academy of Engineering under the Chair in Emerging Technologies programme.}
\thanks{Zhimin Shao is with the Department of Electronic Engineering, Tsinghua University, Beijing, China. She did this work during her internship at National University of Singapore. (\tt\footnotesize{email: shaozhimin00@126.com)}}
\thanks{Jialang Xu, Danail Stoyanov, and Evangelos Mazomenos are with the Wellcome/EPSRC Centre for Interventional and Surgical Sciences and the
Department of Medical Physics and Biomedical Engineering, University College London, London, UK. (\tt\footnotesize{email: \{jialang.xu.22; danail.stoyanov; e.mazomenos\}@ucl.ac.uk})}%
\thanks{Yueming Jin is with the Department of Biomedical Engineering and the Department of Electrical and Computer Engineering, National University of Singapore, Singapore. (\tt\footnotesize{email: ymjin@nus.edu.sg)}}
\thanks{Zhimin Shao and Jialang Xu contributed equally. Joint senior and corresponding authors: Yueming Jin and Evangelos Mazomenos.}
}

\begin{document}
\bstctlcite{IEEEexample:BSTcontrol}

\maketitle
\thispagestyle{empty}
\pagestyle{empty}

\begin{abstract}
Despite significant advancements in robotic systems and surgical data science, ensuring safe and optimal execution in robot-assisted minimally invasive surgery (RMIS) remains a complex challenge. Current surgical error detection methods involve two parts: identifying surgical gestures and then detecting errors within each gesture clip. These methods seldom consider the rich contextual and semantic information inherent in surgical videos, limiting their performance due to reliance on accurate gesture identification.
Motivated by the chain-of-thought prompting in natural language processing, this letter presents a novel and real-time end-to-end error detection framework, Chain-of-Thought (COG) prompting, leveraging contextual information from surgical videos.
This encompasses two reasoning modules designed to mimic the decision-making processes of expert surgeons. Concretely, we first design a Gestural-Visual Reasoning module, which utilizes transformer and attention architectures for gesture prompting, while the second, a Multi-Scale Temporal Reasoning module, employs a multi-stage temporal convolutional network with both slow and fast paths for temporal information extraction. We extensively validate our method on the public benchmark RMIS dataset JIGSAWS. Our method encapsulates the reasoning processes inherent to surgical activities enabling it to outperform the state-of-the-art by 4.6\% in F1 score, 4.6\% in Accuracy, and 5.9\% in Jaccard index while processing each frame in 6.69 milliseconds on average, demonstrating the great potential of our approach in enhancing the safety and efficacy of RMIS procedures and surgical education. The code will be available.

\end{abstract}

\section{INTRODUCTION}
The advent of robot-assisted minimally invasive surgery (RMIS) has revolutionized operative procedures across various medical specialties, from urology to general surgery. RMIS extends human dexterity, offering unprecedented precise instrument navigation and enabling vivid observation of surgical scene~\cite{maier2017surgical,maier2022surgical,d2021accelerating}. 
Despite clear advancements, RMIS requires a high level of proficiency for surgeons to master the manipulation of sophisticated robotic systems. The safety of RMIS can be inevitably compromised due to technical errors~\cite{JOICE1998409, elhage2015assessment}, such as unintended instrument operation, alteration of the surgeon's intent, and unresponsive robotic systems. Approximately 10-15\% of surgical patients in the UK experience adverse events, of which 50\% are preventable~\cite{vincent2001adverse, healey2002complications}, while 10,624 adverse events in robotic procedures were reported in the US from 2000 to 2013~\cite{collins2019utilising}. 
Technical errors during surgery have become a leading cause of postoperative complications, resulting in reoperations and readmissions~\cite{regenbogen2007patterns}. A lack of standardized RMIS training is identified as one of the main reasons for intraoperative risk to patients \cite{collins2020training}.

In this context, real-time surgical error detection becomes paramount in mitigating the risks associated with RMIS~\cite{liu2022real, aspart2022clipassistnet}. 
By providing immediate feedback to surgeons during live surgeries, real-time error detection mechanisms can alert surgeons about potential adverse events, and allow for immediate remedy actions to avoid complications. In surgical training and education, real-time error detection can assist trainees in immediately recognizing and correcting their mistakes by pinpointing areas for improvement, thereby accelerating the learning curve and education efficacy~\cite{schreuder2012training}. Furthermore, error detection contributes to more detailed surgical skill assessment. According to~\cite{anastasiou2023keep}, fluctuations in the Global Rating Scale during surgery indicate suboptimal performance, with significant deviations suggesting the occurrence of human errors. Once identified, these can serve as valuable indicators of surgical proficiency~\cite{morita2020real}.

However, real-time error detection poses significant challenges due to the complicated nature of surgical procedures and the human involvement in operating surgical robots. For instance, although repeated attempts are regarded as errors, certain actions may be repeated intentionally by the surgeon to achieve the desired outcome. Hutchinson et al.~\cite{hutchinson2022analysis} have conceptualized the surgical context as a hierarchical structure, ranging from the overall procedure down to the specific gesture and motion. They have annotated the open-source JHU-ISI Gesture and Skill Assessment Working Set (JIGSAWS)~\cite{gao2014jhu} with error labels through human inspection of videos and the available gesture labels and also introduced a framework for assessing both executional and procedural mistakes by analyzing kinematic data. The findings illustrate that error types and frequencies significantly differ in various tasks and gestures (e.g., pulling a suture, or passing needles). 

Recent advances in surgical error detection have led to the development of two separate parts: gesture recognition and error detection within each gesture type~\cite{yasar2020real}.
Early works utilized conventional deep learning techniques, e.g., convolutional neural networks (CNNs) and long short-term memory (LSTM), to predict potential unsafe events caused by unintentional human errors in simulated surgical training tasks~\cite{yasar2020real, li2022runtime} and retinal microsurgery~\cite{he2019enabling}. Li et al.~\cite{li2022runtime} designed a Siamese CNN-LSTM network to contrast the trajectories of normal and erroneous gestures and improved the error detection for each type of gesture to an online mode, reporting state-of-the-art performance on JIGSAWS. However, most of these methods rely heavily on the gesture label as prior knowledge to segment the surgical video. Generally, the overall performance of error detection in the two-part framework depends on external expertise to define gestures in advance, or on the results from gesture recognition that initially segments and categorizes a surgical video into gesture clips. This motivates us to explore the end-to-end error detection method to avoid gesture annotation.

Another limitation of prior work is ignoring contextual and semantic information in surgery, since they focus on analyzing fragments of kinematic data instead of the whole surgical video~\cite{zia2018automated, ismail2019accurate,van2021gesture}.  Although kinematic data is informative, it captures only the dynamics of surgical tools. Surgical errors are also expected to be influenced by factors like the surgical tools employed and their interactions with anatomical structures within the surgical workspace~\cite{funke2019using}, aspects that the endoscopic video can capture directly. Besides, the error duration has high variation and errors occur in different time scales. Error types like wrong gesture procedures and multiple attempts usually occur across actions and therefore last a long time while the error out of view occurs within the gesture, thus lasting short. This semantic information is also contained in videos and not present in kinematic data~\cite{ZAPPELLA2013732}. While video-based methods have been extensively validated to achieve comparable or even superior performance to kinematic-based methods in other tasks within surgical data science, such as skill assessment~\cite{funke2019video, wang2020towards, anastasiou2023keep}, gesture recognition~\cite{ZAPPELLA2013732} and instrument segmentation~\cite{jin2019incorporating}, the effectiveness of video-based approaches that capture short and long term temporal information in error detection remains unexplored.

\begin{figure}
\centerline{\includegraphics[width=\columnwidth]{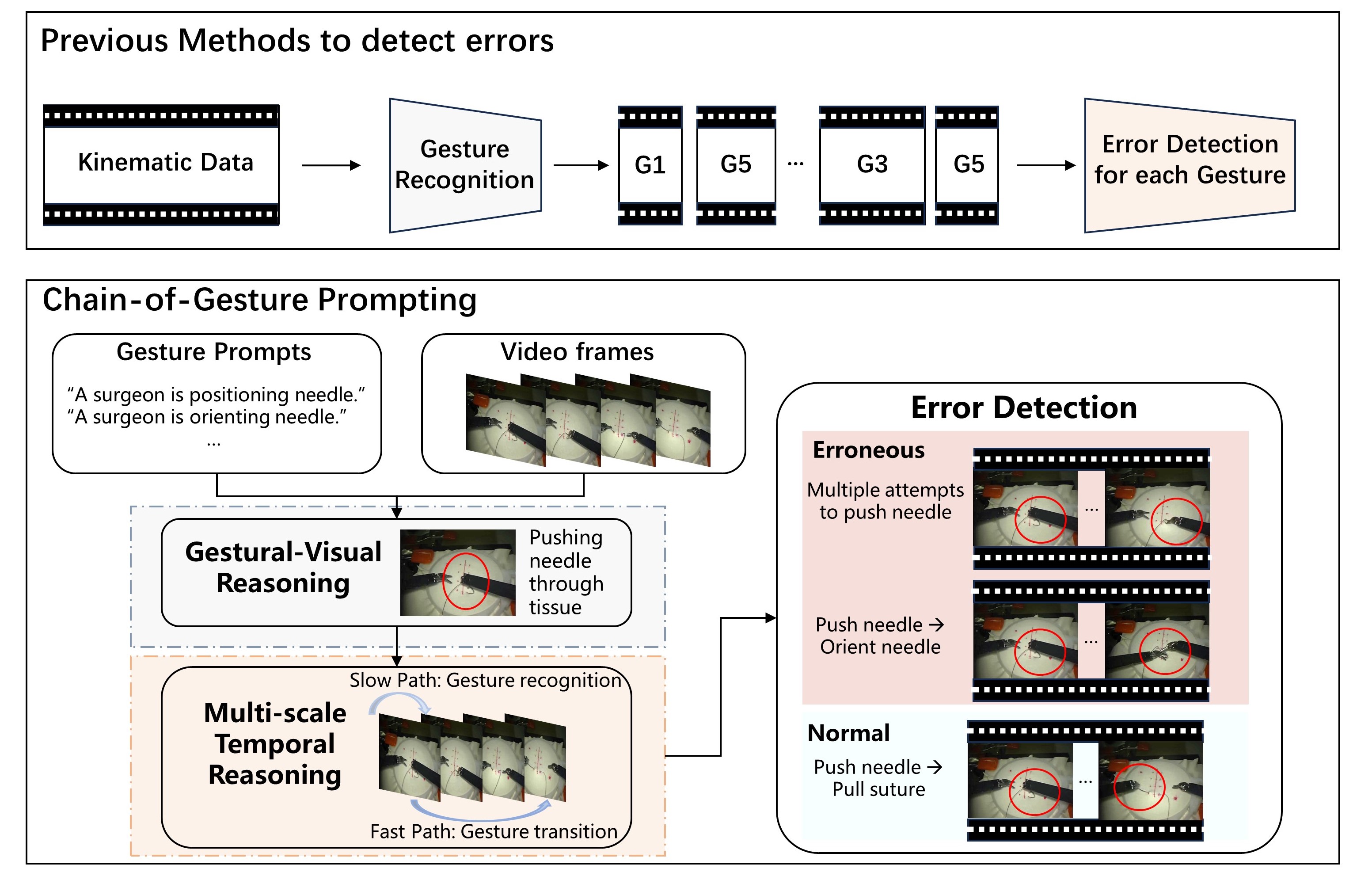}}
\caption{Illustration on previous methods and our proposed Chain-of-Gesture prompting. (a) Previous methods detect errors with two separate parts: gesture recognition and error detection for each type of gesture. (b) We propose an end-to-end Chain-of-Gesture prompting framework to capture complex visual reasoning processes with two reasoning modules: Gestural-Visual reasoning and Multi-scale Temporal Reasoning.}
\label{fig1}
\end{figure}

In this paper, we propose a novel Chain-of-Gesture (COG) prompting framework for real-time surgical error detection from robotic surgical videos.
Inspired by chain-of-thought~\cite{wei2022chain} prompting used in natural language processing that divides a problem into a sequence of intermediate reasoning steps, our COG model consists of sequential reasoning steps to streamline the error detection process.
Our approach involves two reasoning modules that mimic the two parts of gesture recognition and error detection within each gesture type, ultimately forming an integrated end-to-end error detection framework, as shown in Fig.~\ref{fig1}. 
Concretely, we first propose Gestural-Visual Reasoning (GVR), which uses language prompts with attention, to locate gestures in videos. We use vision-language models to generate language prompts for a predefined set of gestures, serving as potential gestural information, and then augment the video with additional informative gesture cues without extra annotation cost through a transformer layer and an attention layer. Based on the augmented features, we further develop the Multi-Scale Temporal Reasoning (MSTR) to capture both slow and fast temporal transitions. It is achieved by two temporal feature extraction paths in different time scales and prediction consistency loss across multi-scales. We extensively evaluate our method on the JIGSAWS dataset. Our method outperforms existing state-of-the-art approaches significantly. The main contributions of the paper are as follows:

\begin{itemize}
    \item We introduce an end-to-end surgical error detection framework that operates in real-time by chain-of-gesture prompting without needing two separate parts or additional gesture labels.
    \item Our model incorporates gesture clues with videos through GVR and analyzes surgical procedures at both fine and coarse temporal scales through MSTR with 1.7\% and 2.2\% improvement in F1 score respectively, enhancing the fine-grained and overall understanding of the surgical context.
    \item Our model achieves remarkable improvements compared with state-of-the-art methods for surgical error detection.
\end{itemize}

\section{METHODS}

\begin{figure*}
\centerline{\includegraphics[width=2\columnwidth]{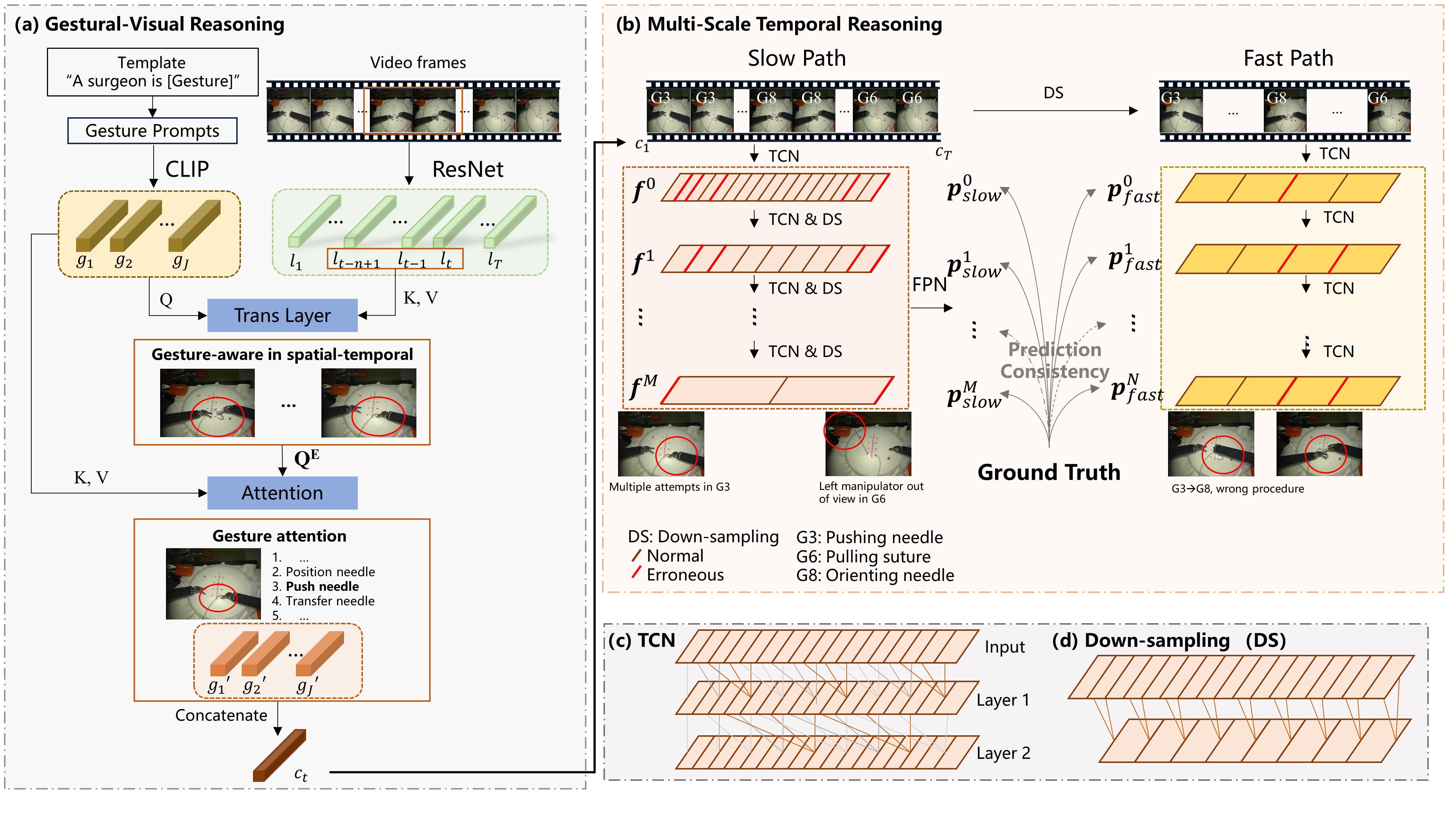}}
\vspace{-2mm}
\caption{Overview of our proposed Chain-of-Gesture. (a) Gestural-visual reasoning module with gesture prompts and visual embedding, a transformer layer, and an attention layer for gestural prompting. (b) Multi-scale temporal reasoning module with a slow path and a fast path is optimized by prediction consistency. (c) Temporal Convolutional Network (TCN) in detail. (d) Downsampling in detail.}
\label{fig2}
\end{figure*}

The overview of our proposed COG is illustrated in Fig.~\ref{fig2}. In this section, we first present the problem formulation for the surgical error detection task. Then we detail the design and functions of GVR and MSTR modules in the proposed COG, which are tailored for integrating gestural and temporal information. Finally, the prediction consistency loss across multi-scales is used to improve the accuracy and consistency of predictions across both individual video frames and their adjacent segments.

\subsection{Problem Formulation}

During surgical procedures, both human surgeons and robotic systems are susceptible to making mistakes that can adversely affect the outcome. Surgical errors are broadly categorized into two types: executional and procedural errors~\cite{hutchinson2022analysis}. Executional errors encompass issues such as repeated attempts at a specific action or the misplacement of instruments beyond the endoscopic view. Procedural errors, on the other hand, refer to the omission or incorrect sequencing of gestures that are correctly performed in isolation. Unlike prior studies~\cite{yasar2020real,li2022runtime} that primarily employ kinematic data analysis within predefined temporal windows to detect executional errors for every type of gesture, our study aims to identify errors, irrespective of their type, in real-time by analyzing surgical videos without the need for gesture labels. 

Given a surgical video dataset $\mathcal{(X, Y)}$, where $\mathcal{X}$ consists of the video frames and $\mathcal{Y}$ denotes the error labels, categorizing each frame as either erroneous or normal. Our model processes a continuous video stream, $X_t = \{x_i\}_{i=1}^t$, to identify errors $y_t$ in the current frame $x_t$, in a real-time detection mode. Here, $x_i$ refers to the $i$-th video frame within the sequence $X_t$.

\subsection{Gestural-Visual Reasoning (GVR)}

Surgical procedures can be conceptualized as hierarchical structures, organized into a sequence of tasks that consist of discrete steps, known as gestures~\cite{hutchinson2022analysis}. These gestures represent fundamental surgical actions. Within this context, errors are closely associated with the gestures performed by surgeons, influencing the detection of errors throughout the procedure. It is essential to recognize the distinct characteristics of various surgical scenarios, as the frequency and type of errors vary significantly across different gestures involved in surgical tasks. For instance, when pushing the needle through tissue, trainees often encounter multiple attempt errors. This issue primarily arises from difficulties in needle handling, including incorrect angle or depth of entry, which requires repeated efforts to correctly position the needle. On the other hand, pulling the suture tends to result in out-of-view errors. Therefore, modeling the dynamics of gestures within the visual domain of endoscopic videos offers a crucial semantic understanding of surgical scenes.

To achieve this goal, we propose a GVR module by structuring the visual features of video frames around a predefined set of gesture prompt features, as shown in Fig.~\ref{fig2} (a). Firstly, we employ a gesture prompt template $t(\cdot)$: "A surgeon is [Gesture] in the surgery", where [Gesture] is one of $J$ predefined gestures. Subsequently, we generate gesture prompt feature set $\{g_j\}_{j=1}^J\in R^{J\times d_{text}}$ utilizing the CLIP text encoder~\cite{radford2021learning}, where $d_{text}$ is the dimension of textual embedding.
\begin{equation}
    g_j =  CLIP(t(\text{Gesture}_j)), j\in[1,J]
    \label{eq1}
\end{equation}

For the analysis of spatial information, the current video frame $x_t$ is forwarded to a standard CNN model (ResNet50~\cite{he2016deep} in our work) to extract discriminative spatial embeddings $l_t\in R^{2048}$. Subsequently, we enhance the gesture prompt features with spatial awareness by introducing a Transformer layer coupled with an Attention layer~\cite{vaswani2017attention}. 
The transformer layer identifies which frame or parts of frames are most similar to the current gesture prompt by scanning through the last $n$ frames and measuring the similarity between gesture prompts and spatial embeddings derived from video frames. Then, the attention layer complements the transformer by focusing specifically on matching all $n$ recent video frames to the most relevant gesture prompt to obtain spatial-aware gesture prompt features.

Specifically, the transformer layer employs a multi-head attention mechanism, where the gesture prompt feature set $\{g_j\}_{j=1}^J$ act as the query, and the sequence of visual features $l_{t-n+1:t}$  with length $n$ serves as both key and value, which is sequentially followed by a layer normalization, a feed-forward layer, and another layer normalization~\cite{vaswani2017attention}, outputting the refined gesture prompt features $Q^E\in R^{J\times d}$, integrating both gestural and visual cues from the recent video frames:
\begin{equation}
    Q^E = \text{Trans}\left(\{g_j\}_{j=1}^J, l_{t-n+1:t}, l_{t-n+1:t}\right)
    \label{eq2}
\end{equation}

The attention layer utilizes the scaled dot-product attention mechanism:
\begin{equation}
   \text{Atten}(Q,K,V)=\text{softmax}(\frac{QK^T}{\sqrt{d}})V
    \label{eq3}
\end{equation}
where $Q, K, V, \sqrt{d}$ denote query, key, value, and scaling factor respectively. The attention layer takes $Q^E$ as the query and the original gesture prompt feature set $\{g_j\}_{j=1}^J$ as both the key and value to produce spatial-aware gesture prompt features $g_j^\prime \in R^{d}$. This mechanism allows for precisely adjusting the attention weights, ensuring that the model focuses on the most relevant spatial-aware features for each gesture prompt.
\begin{equation}
g_j^\prime=\text{Atten}\left(Q^E, \{g_j\}_{j=1}^J, \{g_j\}_{j=1}^J\right), \ \ j\in[1,J]
\label{eq4}
\end{equation}

For comprehensive spatial-aware gesture representation, we concatenate all $J$ spatial-aware gesture prompt features, generating a cohesive feature $c_t \in R^{Jd}$ for the current frame $x_t$. This aggregation enhances the model's ability to interpret and reason about the spatial dynamics of each gesture within the surgical video.

\subsection{Multi-Scale Temporal Reasoning (MSTR)}
According to their definition, executional errors typically occur within fine temporal scales, whereas procedural errors happen over broader temporal scales~\cite{hutchinson2022analysis}. Drawing inspiration from the SlowFast architecture for video recognition~\cite{feichtenhofer2019slowfast}, we introduce an MSTR module to address the complexities of temporal dynamics in surgical video analysis, as shown in Fig.\ref{fig2} (b). MSTR is bifurcated into two distinct pathways: a Slow Path and a Fast Path to model temporal information at the granular level of individual frames and identify transitions between gestures at the segment level separately. Unlike the SlowFast model~\cite{feichtenhofer2019slowfast}, which reduces the entire video to just two frames, our Fast Path down-samples frame-level cohesive features $c_{1:t}$ by average pooling across every 16 frames to identify transitions between gestures at the segment level, and our Slow Path, which processes the frame-level cohesive features $c_{1:t}$ and coarsens as the stages go deeper. 

Our model encodes temporal cues for both the Slow and Fast Paths by employing a Temporal Convolutional Network (TCN)~\cite{feichtenhofer2019slowfast}. Specifically, the Fast Path incorporates a Multi-Stage TCN (MS-TCN) comprising an initial stage of 10 stacked residual dilated causal 1D convolution layers to generate prediction $\boldsymbol{p}_{\text{fast}}^0$ from the initial stage, followed by $N$ refinement stages, each with 11 causal dilated 1D convolution layers, to obtain refined predictions $\{\boldsymbol{p}_{\text{fast}}^i\}_{i=1}^N$. The configuration of TCN is illustrated in Fig.~\ref{fig2} (c) as an example of a 2-layer TCN. For each layer in TCN, the operation can be formulated by
\begin{equation}
\begin{aligned}
    Z_l &= \text{ReLU}(W_{1,l} * F_{l-1} + b_{1,l})\\
    F_l &= F_{l-1} + W_{2,l} * Z_{l} + b_{2,l}
\end{aligned}
\label{eq5}
\end{equation}
where $F_{l}$ is the output of the layer $l$, $*$ denotes the convolution operator, $W_{1,l}$ is the causal dilated 1D convolution kernel~\cite{farha2019ms}, $W_{2,l}$ is the weight of a 1D convolution and $b_{1,l}$, $b_{2,l}$ are bias vectors.

The Slow Path utilizes a similar MS-TCN architecture but differs by applying MS-TCN to features directly, rather than predictions. Firstly, we employ a TCN to generate the initial feature $\boldsymbol{f}^0$ as the first stage. As the network delves into deeper stages, this path systematically down-samples the temporal resolution by using average pooling with both kernel size and stride set to $k$ (shown in Fig.~\ref{fig2} (d)) at each stage, aiming to compress temporal information effectively. Subsequently, a Feature Pyramid Network (FPN)~\cite{lin2017feature} aggregates features of varying scales $\{\boldsymbol{f}^i\}_{i=0}^M$ to synthesize multiple predictions $\{\boldsymbol{p}_{slow}^i\}_{i=0}^M$. We take $\boldsymbol{p}_{slow}^0$ as the final frame-level prediction.
\begin{equation}
    \boldsymbol{f}^0 = \text{TCN}(c_{1:t})
    \label{eq6}
\end{equation}
\begin{equation}
     \boldsymbol{f}^i = \text{AvgPool}\left(\text{TCN}(\boldsymbol{f}^{i-1})\right)
     \label{eq7}
\end{equation}
\begin{equation}
    \{\boldsymbol{p}_{slow}^i\}_{i=0}^M = \text{FPN}(\{\boldsymbol{f}^i\}_{i=0}^M)
    \label{eq8}
\end{equation}
where $\text{TCN}(\cdot)$ means a single-stage TCN.

\subsection{Prediction Consistency across Multi Scales}
For a video comprising a total of $T$ frames, the prediction length produced by Fast Path becomes $\lfloor T/16\rfloor$ after down-sampling. In Slow Path, the temporal length of predictions at each stage is determined by $T^{i+1} = \lfloor T^i /k\rfloor$, where $T^i$ represents the temporal length of the $i$-th stage. Therefore, predictions from different stages are in various time scales. SlowFast~\cite{feichtenhofer2019slowfast} and SF-TMN~\cite{zhang2023sf}, researchers combine the predictions from slow and fast paths to generate final predictions to merge coarse and fine temporal information. However, the effectiveness is hindered by constraints in prediction accuracy. In contrast, inspired by~\cite{ding2022exploring}, we ensure prediction consistency by adapting the ground truth $y_t$ to align with the temporal resolution of each stage in both the Slow Path and Fast Path through down-sampling. Subsequently, the losses across all stages in two paths are aggregated to compute the total loss. Within each stage, the loss is composed of two parts: a Cross-Entropy (CE) loss calculated at each time point to assess the accuracy of predictions, and a Mean Squared Error (MSE) calculated over the detection probabilities between every two adjacent time points to ensure smooth transitions in the prediction sequence.
\begin{equation}
\begin{aligned}
    \mathcal{L}_{CE} &= \mathcal{L}_{CE-slow} + \mathcal{L}_{CE-fast}\\
    &=-\frac{1}{M+1}\frac{1}{T^{i}}\sum_{i=0}^{M}\sum_{t=1}^{T^{i}}y_{t}^i\log (p_{slow,t}^i) 
    \\ & \ \ -\frac{1}{N+1}\frac{1}{\lfloor 
    T/16\rfloor}\sum_{j=0}^{N}\sum_{t=1}^{\lfloor T/16\rfloor}y_{t}^j\log (p_{fast,t}^j)
\end{aligned}
\label{eq9}
\end{equation}
\begin{equation}
\begin{aligned}
    \mathcal{L}_{MSE} &= \mathcal{L}_{MSE-slow} + \mathcal{L}_{MSE-fast}\\
    &=\frac{1}{M+1}\frac{1}{T^{i}}\sum_{i=0}^{M}\sum_{t=1}^{T^{i}}|p_{slow,t}^i-p_{slow,t-1}^i|^2 +
    \\ & \ \ \frac{1}{N+1}\frac{1}{\lfloor 
    T/16\rfloor}\sum_{j=0}^{N}\sum_{t=1}^{\lfloor T/16\rfloor}|p_{fast,t}^j-p_{fast,t-1}^j|^2
\end{aligned}
\label{eq10}
\end{equation}
where $y_{t}^i$ and $y_{t}^j$ are the corresponding ground truth at $i$-th stage in Slow Path and $j$-th stage in Fast Path, respectively. 

Hence, the overall objective of our COG is
\begin{equation}
\mathcal{L}_{total}= \mathcal{L}_{CE}+\lambda \mathcal{L}_{MSE}
\label{eq11}
\end{equation}

\section{Experiments}
\subsection{Datasets and Evaluation Metrics}
JIGSAWS~\cite{gao2014jhu} is a public dataset derived from the \textit{da Vinci} surgical system, capturing data from eight surgeons performing three dry lab surgical tasks: suturing, knot tying, and needle passing. The dataset includes synchronized kinematic and 640$\times$480 resolution video data, recorded at 30Hz, alongside manual annotations. These annotations detail gestures associated with each trajectory and performance scores reflecting surgeon proficiency. Hutchinson et al~\cite{hutchinson2022analysis} extends JIGSAWS by annotating executional and procedural errors at the frame level for suturing and needle-passing tasks. In our research, we utilize the video data and all corresponding error labels, irrespective of error type, with each video frame being categorized as either normal (0) or erroneous (1). Note that we only use the description of potential gestures from the gesture vocabulary provided by~\cite{gao2014jhu}, rather than the detailed gesture ground truth of each frame.

To evaluate our method, we employ the Leave-One-Supertrial-Out (LOSO) cross-validation, as detailed in the original JIGSAWS paper~\cite{gao2014jhu}. All surgeons repeated each surgical task five times. In LOSO, the $i$-th trial of each surgeon is excluded from the dataset to serve as the test set, thereby assessing the model's ability to generalize across different trials conducted by the same surgeon. 

This work aims to detect errors in surgical video in real-time. Therefore, we employ the F1 score, Accuracy, and Jaccard index at the frame level to evaluate the performance. Furthermore, to ensure an explicit and fair comparison with the state-of-the-art work on surgical error detection~\cite{li2022runtime}, we follow its evaluation protocol and also use a 2-second sliding window with a 1.2-second stride to generate window-level results. The ground truth for each trial out under the LOSO setting is detailed in Table~\ref{tab1}. Given a minor imbalance between categories, the F1 score was chosen as the main evaluation metric.

\begin{table}
\begin{center}
\caption{The number of frames and windows for each trial in LOSO.}
\label{tab1}
\begin{tabular}{c|cc|cc}
\hline
             & \multicolumn{2}{c|}{\#frames} & \multicolumn{2}{c}{\#windows} \\ \hline
LOSO setting & Total     & Erroneous       & Total      & Erroneous      \\ \hline
Trial 1 out      & 8332      & 5453 (65\%)     & 1076       & 764 (71\%)      \\
Trial 2 out      & 6056      & 3486 (58\%)     & 775        & 466 (60\%)     \\
Trial 3 out      & 7066      & 3739 (53\%)     & 886        & 500 (56\%)     \\
Trial 4 out      & 6979      & 3277 (47\%)     & 868        & 448 (52\%)     \\
Trial 5 out      & 5433      & 2353 (43\%)     & 640        & 308 (48\%)     \\ \hline
\end{tabular}
\end{center}
\end{table}

\begin{table*}[]
\begin{center}
\caption{Quantitative results on frame level and window level of comparisons between SOTA methods and our proposed Chain-of-Gesture model on JIGSAWS dataset. $\Delta$ GVR: no Gestural-Visual Reasoning; $\Delta$ MSTR: no Multi-Scale Temporal Reasoning; $\Delta$ Slow Path: no Slow Path module; $\Delta$ Fast Path: no Fast Path module. \textbf{K}: kinematic data, \textbf{V}: video data. * denotes methods focusing on temporal information extraction.}
\label{tab2}
\begin{tabular}{c|l|ccc|cccc}
\hline
\multirow{2}{*}{Input} & \multirow{2}{*}{Method} & \multicolumn{3}{c|}{Frame level}                          & \multicolumn{4}{c}{Window   level}  \\ \cline{3-9} 
                       &                         & F1                & Accuracy          & Jaccard           & F1                & Accuracy          & Jaccard           & P Values \\ \hline
K                      & CNN-LSTM~\cite{li2022runtime} & --                 & --                 & --                 & 70.0 $\pm$ 1.8          & 65.2 $\pm$ 1.2          & 53.9 $\pm$ 2.2          & 2e-12 \\ \hline
\multirow{10}{*}{V}     & ResNet                  & 70.8 $\pm$ 4.0          & 66.9 $\pm$ 3.3          & 54.9 $\pm$ 4.9          & 72.8 $\pm$ 4.1          & 67.8 $\pm$ 3.1          & 57.4 $\pm$ 5.3          & 5e-09 \\
                       & TeCNO*~\cite{czempiel2020tecno}                   & 69.6 $\pm$ 2.6          & 66.4 $\pm$ 2.4          & 53.4 $\pm$ 3.1          & 71.4 $\pm$ 2.3          & 66.7 $\pm$ 2.4          & 55.5 $\pm$ 2.8          & 6e-12 \\
                       & Trans-SVNet*~\cite{jin2022trans}             & 71.0 $\pm$ 5.5          & 59.3 $\pm$ 9.0          & 55.3 $\pm$ 6.9          & 74.2 $\pm$ 5.3          & 62.6 $\pm$ 8.7          & 59.3 $\pm$ 6.9          & 4e-2 \\
                       & SAHC*~\cite{ding2022exploring}                    & 70.7 $\pm$ 3.5          & 67.8 $\pm$ 2.2          & 54.8 $\pm$ 4.3          & 72.6 $\pm$ 3.1          & 68.0 $\pm$ 2.2          & 57.0 $\pm$ 4.0          & 1e-20 \\
                       & SF-TMN*~\cite{zhang2023sf}                  & 69.8 $\pm$ 3.4          & 66.9 $\pm$ 2.5          & 53.7 $\pm$ 4.1          & 71.4 $\pm$ 3.4          & 67.1 $\pm$ 2.7          & 55.7 $\pm$ 4.3          & 3e-11
                       \\ \cline{2-9}
                      &
Ours ($\Delta$ GVR)                   & 70.8 $\pm$ 4.2          & 67.0 $\pm$ 3.1          & 54.9 $\pm$ 5.2               & 72.9 $\pm$ 4.4          & 67.9 $\pm$ 3.6          & 57.5 $\pm$ 5.7  & 3e-24     \\ &
Ours ($\Delta$ MSTR)          & 70.0 $\pm$ 4.6          & 64.8 $\pm$ 4.3          & 54.1 $\pm$ 5.7            & 72.4 $\pm$ 4.9          & 66.3 $\pm$ 4.7          & 57.0 $\pm$ 6.3      & 3e-11   \\ &
Ours ($\Delta$ Slow Path)             & 71.3 $\pm$ 4.8          & 64.0 $\pm$ 4.3          & 55.6 $\pm$ 6.0             & 74.0 $\pm$ 4.9          & 66.2 $\pm$ 5.0          & 59.0 $\pm$ 6.5  &  1e-2  \\ &
Ours ($\Delta$ Fast Path)              & 71.2 $\pm$ 3.9          & 66.4 $\pm$ 4.2          & 55.4 $\pm$ 4.9              & 73.6 $\pm$ 4.3          & 67.9 $\pm$ 4.5          & 58.4 $\pm$ 5.6   &7e-3\\   & \textbf{Ours}      & \textbf{72.3 $\pm$ 4.6} & \textbf{68.3 $\pm$ 3.5} & \textbf{56.8 $\pm$ 6.0} & \textbf{74.6 $\pm$ 5.1} & \textbf{69.8 $\pm$ 4.5} & \textbf{59.8 $\pm$ 6.8} & --         \\ \hline
\end{tabular}
\end{center}
\end{table*}

\subsection{Implementation Details}
All experiments are implemented in PyTorch on a single NVIDIA RTX 3060 GPU. For the extraction of spatial embeddings $l_t$, we employed the ResNet-50 model, initially pre-trained on the ImageNet dataset and further fine-tuned on the JIGSAWS dataset with error labels frame-by-frame using Adam optimizer with the learning rate of $1\times10^{-4}$ and a batch size of 64. Video frames are resized into $240\times 240$ and center-cropped to $224 \times 224$. To reduce redundancy and computational demands, video data are downsampled to 5 Hz. For gesture prompt feature $g_j$, we used the pre-trained CLIP ViT-B32~\cite{radford2021learning} model with fixed parameters as our text encoder. The description of gestures is drawn from a common gestural vocabulary \cite{gao2014jhu} comprised of 15 distinct gestures, thus $J=15$. The spatial and gestural embeddings extracted from ResNet-50 and CLIP are used as inputs to our COG model, without further tuning during the COG model's training phase. Our COG is trained end-to-end using the Adam optimizer for 50 epochs with the initial learning rate set to $5\times10^{-4}$. We set $M = N = 3$ empirically and standardize the dimension of all casual dilated 1D convolution layers in the MSTR module to 64. The coefficient of MSE loss $\lambda$ is empirically set to 0.15. The length of sequence $n$ is set to 40, and the kernel size and stride of average pooling $k$ in down-sampling is set to 4.

\subsection{Comparison with State-of-the-Art}
The two existing studies that use CNN-LSTM based on kinematic data for surgical error detection~\cite{yasar2020real,li2022runtime}, report state-of-the-art  F1 scores on JIGSAWS. To ensure a fair comparison, we re-implemented this kinematic-based CNN-LSTM approach and a range of state-of-the-art video-based methods for surgical video analysis. These methods include ResNet-50~\cite{he2016deep}; TeCNO~\cite{czempiel2020tecno}, which employs MS-TCN to capture long-range dependencies in video sequences; Trans-SVNet~\cite{jin2022trans}, which introduces a transformer layer to integrate spatial and temporal features effectively; SAHC~\cite{ding2022exploring}, which explores the use of hierarchical clustering to refine the feature extraction process; and SF-TMN~\cite{zhang2023sf}, which combines features from slow and fast processing paths to enhance motion analysis. All methods were implemented based on their publicly released codes and the methodologies described in the original literature. The comparative results are presented at both the frame-level and window-level in Table~\ref{tab2}.

Comparing different inputs to the model, video data consistently outperforms kinematic data in error detection. Notably, even the spatial information extracted through a fine-tuned ResNet-50 enhances performance significantly, with improvements observed in the window-level metrics: approximately 2.8\% in F1 score, 2.6\% in accuracy, and 3.5\% in Jaccard, further showcasing the importance of spatial information in the accurate detection of errors.

Among the methods that emphasize temporal information extraction (i.e. TeCNO, Trans-SVNet, SAHC, and SF-TMN), Trans-SVNet emerges as a strong contender. It effectively combines spatial and temporal data using a Transformer-based fusion head, making it the second-best method with an impressive 74.2\% F1 score and 59.3\% Jaccard. This achievement indicates that contextual information fusion is effective for the surgical error detection task. Nevertheless, it is crucial to note that the accuracy of Trans-SVNet is considerably lower. This suggests a tendency towards generating false positives which could be attributed to the imbalanced dataset. 

Other methods, despite their sophistication in temporal information extraction, fall short when compared to the more straightforward ResNet-50 approach, possibly due to a misalignment in the temporal scale they employ. While these methods are adept at phase recognition within surgical videos—where the temporal scope is broader and encompasses entire surgical procedures—they appear less suited for the more granular task of error detection. Error detection requires a finer temporal resolution to capture subtle deviations or incorrect actions. Hence, the temporal granularity and feature selection become critical factors in the performance of these models. 

In contrast, our COG introduces gesture prompts for each frame, thereby enriching the data with additional contextual layers, and extracts multi-scale features at fine and coarse temporal resolution, resulting in superior performance across all metrics at both the frame and window levels compared to other methods. COG appears to exploit the spatial-temporal dynamics more effectively. As a result, the COG model not only achieves higher performance metrics but also suggests a methodological advancement in the handling of gestural and contextual information. It also highlights the potential for further refinements in the use of contextual cues to improve the accuracy of such models in complex tasks like surgical error detection. 

In our statistical analysis, we performed the paired T-test between our COG and other methods to calculate the P values for the F1 score at the window level. The COG not only demonstrates a clear enhancement in F1 but also achieves P values significantly below the 0.05 in all compared cases, confirming the robustness and reliability of COG in detecting errors in surgical videos.

\subsection{Ablation Studies}
\subsubsection{Effectiveness of Key Components}
We first analyze the contributions of two distinct reasoning modules (i.e., GVR and MSTR) in our proposed COG model and evaluate the discrete effects of the ``Slow Path'' and ``Fast Path'' in the MSTR, as presented in Table~\ref{tab2}. The omission of either GVR or MSTR led to performance degradation, as evidenced by all metrics. Through statistical analysis, we determined the P values, which quantified the significance of the performance drop. Notably, the absence of GVR had a more profound impact than the exclusion of MSTR, with a smaller P value. This result confirms our hypothesis that the integration of gesture cues via GVR is not merely beneficial but vital to the model's success, as it provides critical contextual information for reasoning about surgical actions. As for the temporal dynamics of MSTR, our findings indicate that the Fast Path plays a more crucial role than the Slow Path in our context of error detection to reflect the gesture transition.
\begin{figure}
\centerline{\includegraphics[width=0.9\columnwidth]{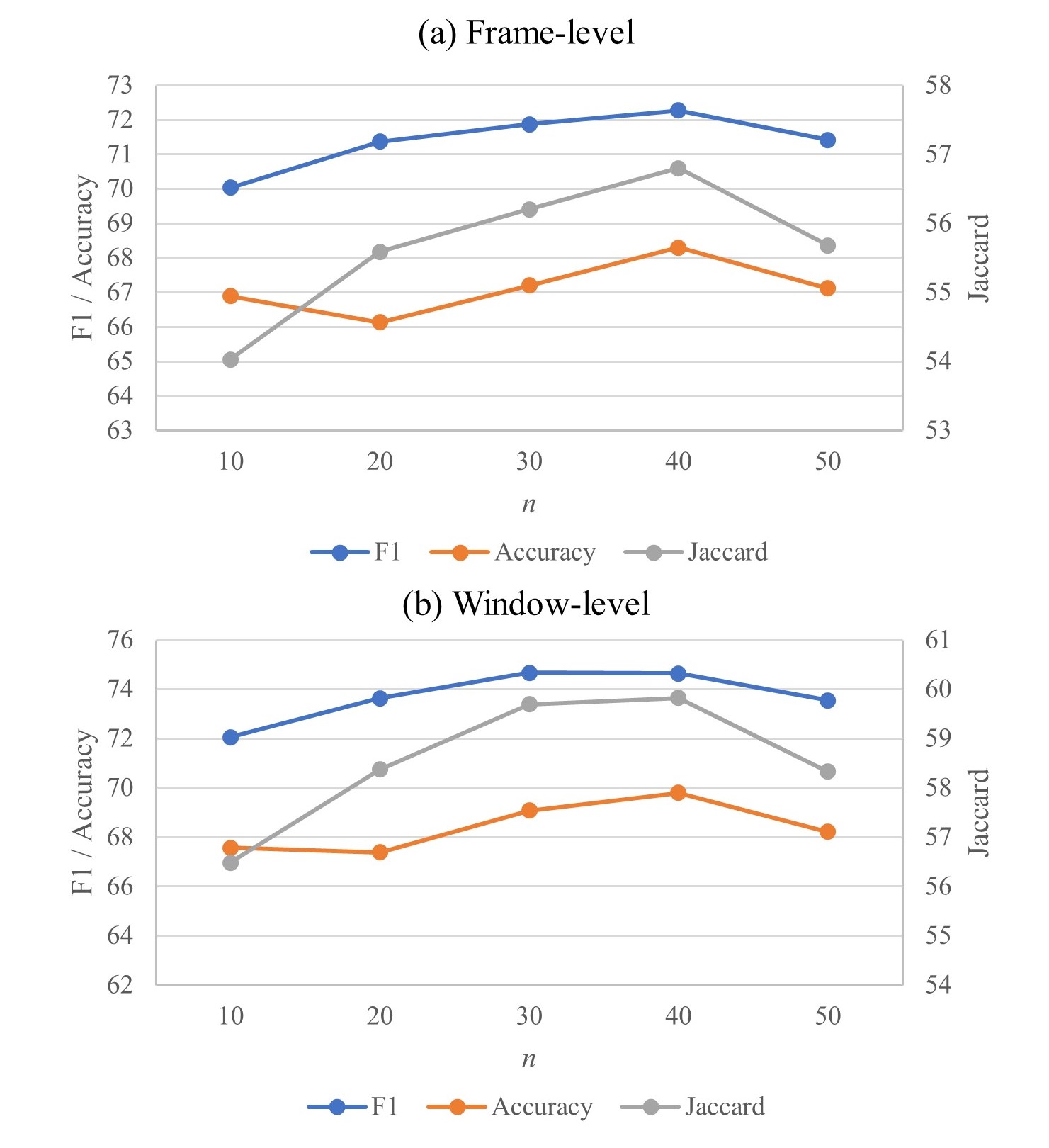}}
\caption{Analysis of length of sequence $n$ used in GVR. We show the results of the F1 score, Accuracy, and Jaccard of models with different $n$.}
\label{fig3}
\end{figure}

\subsubsection{Length of Sequence in GVR}
We then focus on the critical parameter of sequence length, denoted as $n$, which determines the number of video frames considered for identifying its gesture. Figure~\ref{fig3} shows the model performance with different values of $n$ used in Eq.~\ref{eq2}. The empirical findings reveal that a sequence length of $n$=40 frames yields the most promising results. This length corresponds to approximately 1.5 times the average gesture duration in our dataset, which is 27 frames. By extending beyond the average single gesture length, the GVR module is afforded a more holistic view, encompassing not just the gesture itself but also the critical transition phase to the subsequent gesture. This broader temporal window includes the tail end of one gesture and the onset of the next, thus we can discern and identify executional and procedural errors.

\subsubsection{Number of Stages in MSTR}
The number of stages in MSTR determines the temporal dimensions of the predictions from the final stage. Table~\ref{tab3} presents the model performance across different values of $M$ and $N$. We observe that three stages yield the best results; both fewer and greater numbers of stages negatively impact performance. Specifically, when $M=N=2$, the temporal predictions from the slow path are calculated as $\lfloor T /{k^2}\rfloor$. In our experiments, we set $k=4$. Consequently, the temporal predictions from the final stage of both the slow and fast paths are the same, failing to capture longer ranges of information. As for M=N=4, considering the average video length of 538, a four-fold down-sampling with $k = 4$ reduces the average length of the prediction from the final stage in the slow path to only 2, insufficient to encompass the full range of gestures in the videos, thus limiting the effectiveness.

\begin{table}
\begin{center}
\caption{Comparison with different stages $M$ and $N$ in MSTR.}
\label{tab3}
\setlength{\tabcolsep}{1mm}\begin{tabular}{c|ccc|ccc}
\hline
\multirow{2}{*}{M=N} & \multicolumn{3}{c|}{Frame level}                          & \multicolumn{3}{c}{Window level}                          \\ \cline{2-7} 
                     & F1                & Accuracy          & Jaccard           & F1                & Accuracy          & Jaccard           \\ \hline
2                    & 70.9$\pm$4.3          & 67.3$\pm$3.3          & 55.1$\pm$5.3          & 72.6$\pm$4.7          & 67.9$\pm$4.1          & 57.3$\pm$6.1          \\
3                    & \textbf{72.3$\pm$4.6} & \textbf{68.3$\pm$3.5} & \textbf{56.8$\pm$6.0} & \textbf{74.6$\pm$5.1} & \textbf{69.8$\pm$4.5} & \textbf{59.8$\pm$6.8} \\
4                    & 71.4$\pm$4.1          & 65.5$\pm$3.7          & 55.6$\pm$5.1          & 73.5$\pm$3.9          & 66.7$\pm$4.1          & 58.3$\pm$5.1          \\ \hline
\end{tabular}
\end{center}
\end{table}

\subsection{Visual Results}
\begin{figure}
\centerline{\includegraphics[width=\columnwidth]{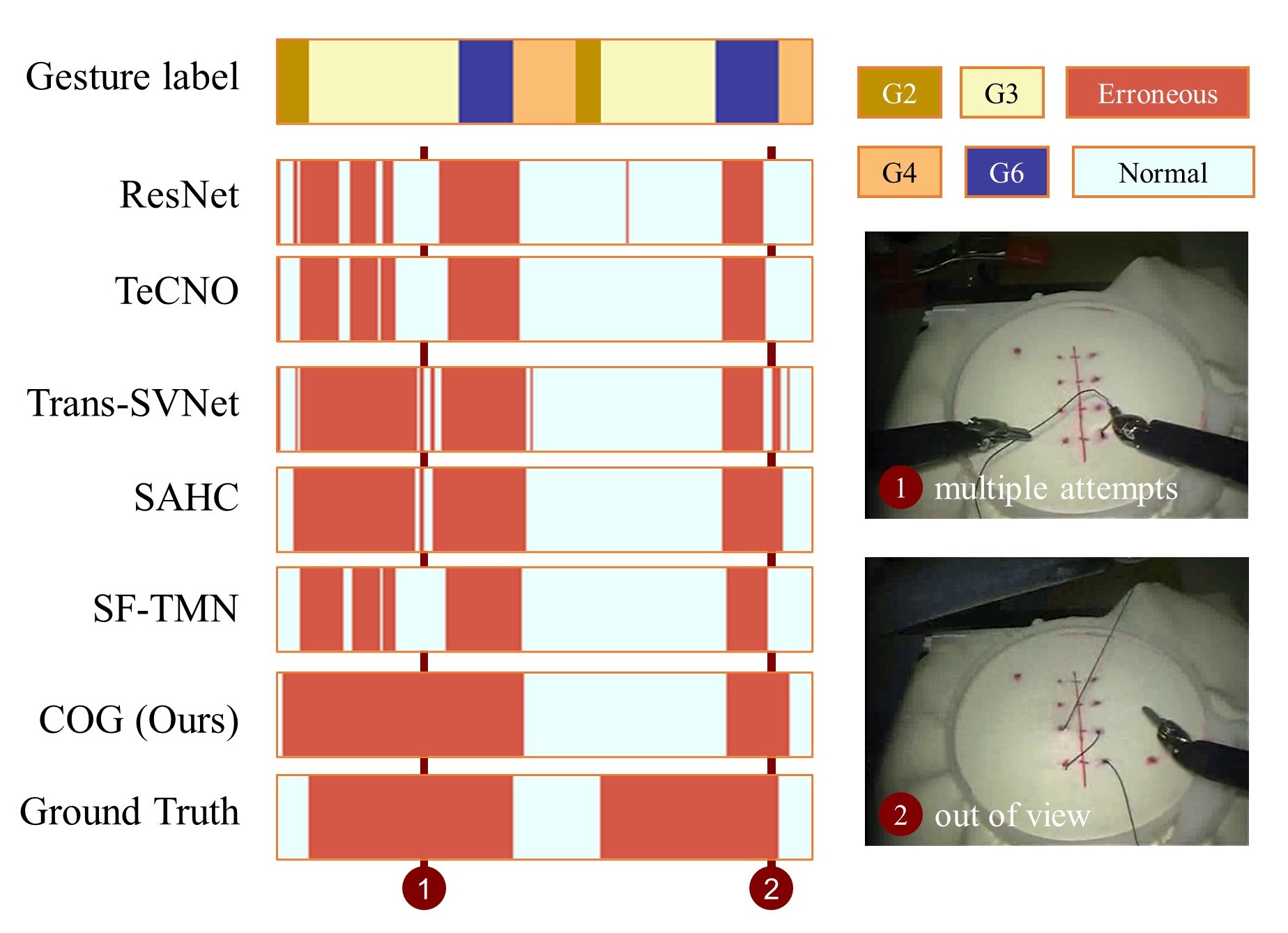}}
\caption{Color-coded ribbon illustration for a suturing video clip.}
\label{fig4}
\end{figure}

In Fig.~\ref{fig4}, we visually represent the error detection outcomes of a suturing video clip. Our model's incorporation of both contextual and temporal information is evident in the consistency and robustness of its predictions. By capturing a wider range of cues over time, the proposed COG model demonstrates an enhanced ability to recognize and flag errors that other methods could miss. Furthermore, we visualize typical errors and results to facilitate an intuitive understanding of the COG model's advantages, such as multiple attempts to push the needle through the tissue and the tool manipulator moving out of the camera's view. More can be found in the supplementary video.

\section{DISCUSSION AND CONCLUSION}
Our study presents a novel COG prompting approach for error detection in RMIS. Different from traditional spatial-temporal feature extraction from kinematic data, this leverages readily available video data and innovates the error detection paradigm by developing two reasoning modules to simulate the expert human decision-making process. Our method incorporates gestural information to effectively discern executional and procedural errors. The proposed COG exhibits a notable 4-6\% improvement in window-level metrics compared to existing methods on the JIGSAWS dataset, significantly without the need for gesture labels in training, a strong prerequisite in previous approaches.

The gestural information inherent in surgical videos encompasses a spectrum of behaviors, such as gesture segmentation, recognition, and transition, making it hard to design a one-size-fits-all method for extracting the critical cues. Current methods typically identify gestures and then detect executional errors within the gesture clip sequentially, reliant on the performance of two distinct parts for gesture recognition and error detection within each type of gesture. In contrast, our COG integrates two reasoning modules as an end-to-end framework. The first module focuses on gesture localization and segmentation, while the second dives into details within gestures and transitions, aiming for a comprehensive understanding of the surgical context. The validity of our approach is supported by ablation studies and visual analysis, reinforcing the logic and efficacy of our chain-of-gesture philosophy to think step by step.

Our COG can be applied in clinical practice, thanks to its computational efficiency (6.69 milliseconds/frame). This efficiency is due to our GVR design, which used the transformer architecture known for its parallel processing capabilities. Moreover, the down-sampling technique in MSTR significantly reduces video length, enhancing information extraction and computational cost-effectiveness. Notably, our COG obviates the need for additional annotation, further streamlining the process. Additionally, our research shows the potential of integrating contextual and temporal data analysis to develop an accurate error detection framework with considerable implications for surgical training. Further work will focus on detecting error types semantically and finding remedial measures, which could significantly enhance the learning curve for novice surgeons.






\bibliographystyle{IEEEtran}
\bibliography{IEEEabrv,root}

\end{document}